\pdfoutput=1

\documentclass[11pt]{article}

\usepackage[preprint]{acl}

\usepackage{times}
\usepackage{latexsym}
\usepackage[T1]{fontenc}
\usepackage[utf8]{inputenc}
\usepackage{microtype}
\usepackage{inconsolata}

\usepackage{amsmath}
\usepackage{amssymb}

\usepackage{graphicx}
\usepackage{subcaption}
\usepackage{float}

\usepackage{booktabs}
\usepackage{threeparttable}
\usepackage{tabularx}
\usepackage{array}
\usepackage{ragged2e}
\usepackage[table]{xcolor}

\renewcommand{\arraystretch}{1.1}

\newcolumntype{P}[1]{>{\RaggedRight}p{#1}}

\usepackage{algorithm}
\usepackage{algorithmic}

\usepackage{listings}
\usepackage{fancyvrb}
\usepackage{comment}

\lstdefinelanguage{json}{
    morestring=[b]",
    stringstyle=\color{teal},
    showstringspaces=false
}

\title{
    An LLM-Agentic Framework for Intelligent
    Work Order Management in Industrial Maintenance
}

\author{
    Nianjun Zhou\thanks{Equal contribution. Corresponding author.} \\
    IBM, Research Division \\
    Yorktown Heights, NY 10598, USA \\
    \texttt{jzhou@us.ibm.com}
    \And
    Dhaval Patel\thanks{Equal contribution.} \\
    IBM, Research Division \\
    Yorktown Heights, NY 10598, USA \\
    \texttt{pateldha@us.ibm.com}
}

\begin{document}

\maketitle


\begin{abstract}

This paper presents an AI-driven framework, \textit{CodeReAct} agent, designed to enhance industrial equipment maintenance through natural language (NL) interactions. The framework employs an advanced AI agentic approach to interpret human instructions, providing knowledge extraction, asset status summarization, and work order recommendations. It identifies patterns and trends and predicts work orders by standardizing industrial data into business objects (BOs) and utilizing analytical function API calls. The AI agent operates through a Thought-Action-Observation process similar to human problem-solving, automating complex analytical tasks typically performed by data scientists. Notably, the Action phase generates executable code, replacing manual coding efforts.

A case study focusing on chiller maintenance demonstrates the framework's capability to analyze multiple years of work orders, and alerts and anomalies are presented. The results indicate that the AI agent framework effectively processes human utterances to analyze data, extract valuable insights, and support informed decision-making, reducing the need for data scientist involvement. By recontextualizing NLP agent frameworks in industrial engineering, the approach enables autonomous function generation and task execution across a range of domain-specific maintenance scenarios, with chillers used as a representative case study.

\end{abstract}

\vspace{-0.5cm}
\section{Introduction}

Industrial asset management is a systematic approach to managing and maintaining assets such as equipment, machinery, and infrastructure to ensure optimal performance, reliability, and longevity. As industrial systems grow more complex, engineers face increasing challenges in maintaining machinery reliability and efficiency. A key concern is understanding the asset's health and accurately predicting maintenance needs, thereby minimizing unplanned downtime and optimizing operational performance~\cite{serradilla2022deep}. Traditional strategies, such as scheduled preventive maintenance, often fall short of these demands, as they may lead to unnecessary downtime or fail to detect emerging issues that could escalate into critical failures~\cite{gulati2009maintenance}. To address these limitations, there is a growing need for data-driven analytics (\cite{balali2020data}) to support condition-based management (\cite{chuang2020condition, goyal2016intelligent}) rather than relying solely on fixed schedules based on expert recommendations.

\begin{figure*}
    \centering
    \includegraphics[width=0.8\textwidth]{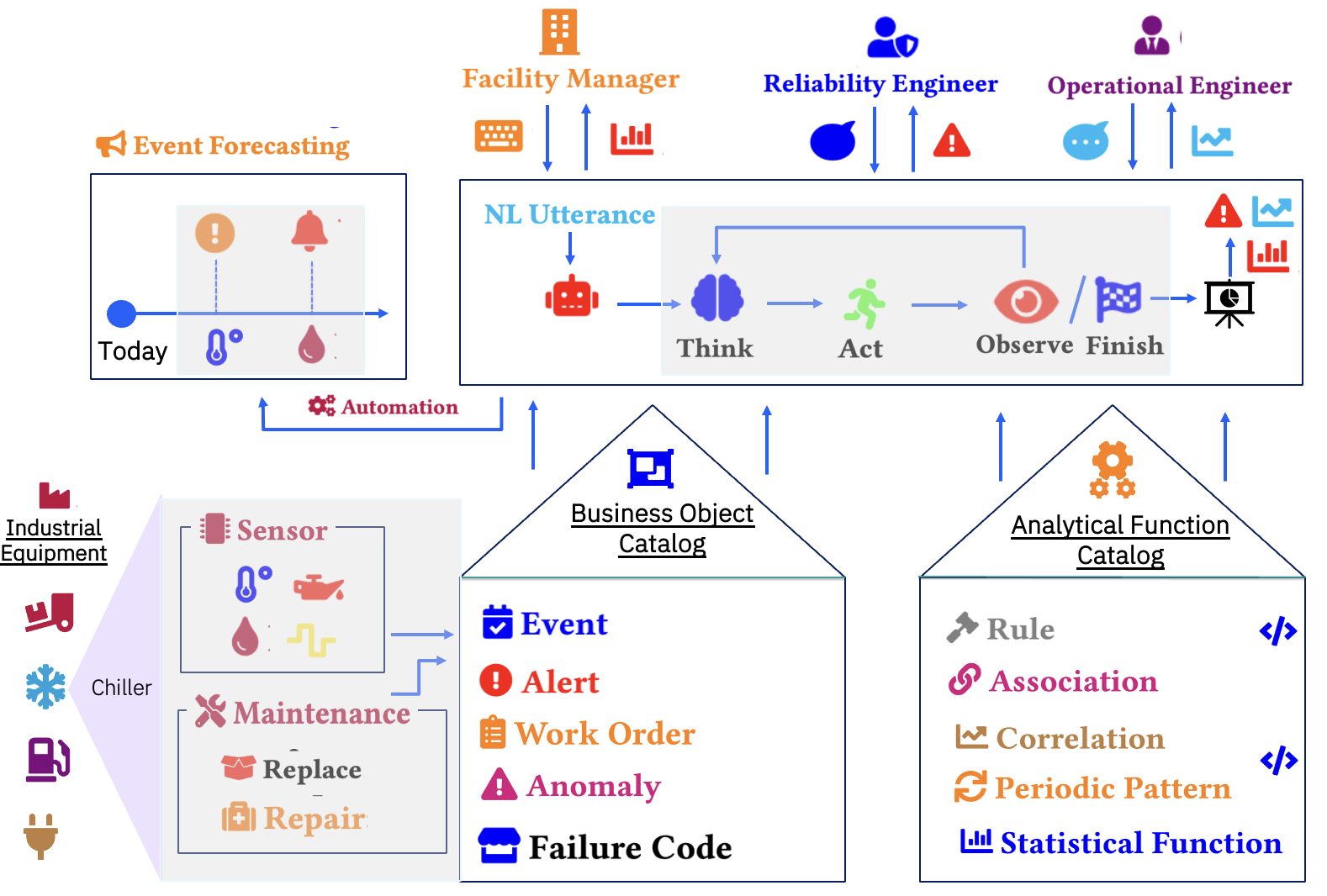}
    \caption{Unified View of the AI Agent-Driven Framework for Event Forecasting and Predictive Maintenance in Industrial Environments}
    \label{fig:intro-motivation}
\end{figure*}

As highlighted in Figure \ref{fig:intro-motivation}, modern \textit{IoT} systems leverage sensor technology to generate vast amounts of data, creating opportunities to enhance predictive maintenance while introducing data integration and interpretation challenges. This data comes from diverse sources, including OEM alerts, AI/ML-generated statistical anomalies~\cite{langbridge2024optimal}, work order requests from site engineers, and subsequent maintenance actions such as repairs or replacements performed by technicians. Integrating both digital and manually generated information presents unique challenges, such as accurately labeling work orders to reflect required maintenance, associating operational alerts and anomalies with specific sensors to derive actionable insights, and linking disparate events to improve decision-making.

Researchers in data mining and data science have developed analytical tools such as frequent/periodic pattern mining~\cite{petsinis2021analysiskeyflavorseventdriven, serradilla2022deep}, association rule discovery, and event prediction~\cite{ye2024miraievaluatingllmagents} to extract engineering and business insights. In industrial asset management, these techniques aid in analyzing maintenance trends, work order variations, and scheduling inefficiencies. For instance, understanding relationships between maintenance tasks involves recognizing recurring patterns, assessing scheduling dependencies, and optimizing task bundling to minimize downtime~\cite{neuberg2003causality}. Additionally, linking alerts and anomalies to corrective actions is crucial for proactive condition-based maintenance~\cite{chuang2020condition, goyal2016intelligent}, enabling early intervention to prevent failures.

Despite the availability of advanced analytical tools~\cite{mcguire2013making}, effectively utilizing them remains challenging—particularly when combining multiple tools for tasks beyond their original scope. Building adaptive systems that integrate new functions and handle evolving data often requires dedicated data science teams due to limitations in flexibility and scalability~\cite{tawil2024trends}. Recently, LLM-supported AI agents have begun transforming operations by automating workflows, such as SWE-Agent for GitHub issue resolution~\cite{tao2024magis, jimenez2023swe, la2024can}. In industrial settings, LLM agents can support engineers by answering natural language queries (e.g., “How does my asset pump operate?”) and performing repetitive analysis across domains. However, challenges persist, including hallucinations, tool misuse, domain-specific language barriers, and lack of analytical grounding. Addressing these limitations is key to recontextualizing NLP agents for robust, cross-domain decision-making in industrial environments.


\subsection{AI Agent in Industrial Setting}
The framework illustrated in Figure~\ref{fig:intro-motivation} presents a proposed AI agent-driven system for industrial environments. The upper section of the diagram highlights key stakeholders—Facility Manager, Reliability Engineer, and Operational Engineer—who interact with the AI agent (depicted as a robot icon) via natural language (NL Utterance). The agent operates within a ``Think, Act, Observe, Finish'' cycle, mirroring human problem-solving while leveraging automation to enhance decision-making and streamline processes. Among the various tasks the AI agent performs, the ``Event Forecasting'' module is central to the system, represented by a time-series graph and a bell curve symbolizing predictions (e.g., ``Today 0°'').

The bottom partition of the diagram illustrates the data flow and highlights two critical building blocks of the system. Sensor data from industrial equipment (e.g., chillers) triggers events, alerts, and maintenance actions, categorized into a ``Business Object Catalog'' (Events, Work Orders, Alerts, Anomalies, Failure Codes, and their mappings). Business Objects (BOs) define domain entities and enable interoperability between analytic functions. Due to the complexity of these entities, it is challenging for LLMs to comprehend them fully without structured information. However, structured BOs improve message exchange with the AI agent, enhancing data interpretation and insights. The AI agent processes this data through the ``Analytical Function Catalog,'' which includes domain-specific techniques such as rule-based analysis, event correlation, pattern recognition, statistical analysis, and forecasting. This processed information supports the agent's decision-making functions and enables actionable insights that optimize maintenance strategies and operational efficiency.

In this research, we develop and validate the AI agent framework, referred to as a \textit{CodeReAct} agent, through a case study in chiller maintenance. While chillers are used as a representative example, the framework generalizes across asset types such as AHUs, pumps, and generators. Our key contributions are summarized as follows: 1. \textbf{Agentic Workflow Perspective}: We extend the \textit{ReAct} framework (Thought-Action-Observation) with coding capability to support asset-intensive analytics. The resulting agent architecture is domain-agnostic and applicable to diverse industrial scenarios requiring extensive data processing; 2. \textbf{Application Perspective}: By structuring industrial data into \textbf{Business Objects} (BOs) and applying event-driven \textbf{Analytical Functions}, we build a system that allows stakeholders to directly query operational insights without continuous data science support. The framework supports online deployment for periodic monitoring and proactive alerting and recommendations; 3. \textbf{Algorithmic Perspective}: To enhance the agent's performance, we introduce an external reflection mechanism that integrates feedback from execution failures, enabling dynamic system prompt refinement. Additionally, an adaptive model temperature strategy improves task completion and system reliability.

\section{Related Work}

\subsection{AI Agents and Action-Oriented Frameworks}
The evolution of AI agents has seen significant advancements, particularly with frameworks like ReAct (\cite{Yao2022ReActSR}), which introduced reasoning-then-action strategies for structured decision-making. However, traditional approaches often rely on static output formats, such as JSON or predefined text, which limit flexibility in multi-tool integration. Similarly, LangChain has facilitated tool-based solutions for natural language processing tasks but struggles with dynamic task management in complex environments.

The \textbf{CodeAct framework} (\cite{wang2024executable}) advances these approaches by enabling Large Language Models (LLMs) to generate and execute Python code as actions. This innovation allows for seamless integration with Python's ecosystem, enabling dynamic workflow adjustments and real-time multi-tool reasoning. CodeAct enhances adaptability and precision, making it especially effective in industrial applications such as predictive maintenance, where flexibility and task automation are critical.

In addition, \textbf{OpenHands} (\cite{wang2024openhands}), provide an open platform for AI software agents that interact with digital tools in a more generalized and adaptable manner. OpenHands enables AI-driven workflows that involve web interaction, code execution, and command-line operations, extending the capabilities of existing agent-based frameworks.

\subsection{Industrial Asset Event Analytics}

Langbridge et al. (2024) propose a novel approach leveraging Optimal Transport for efficient and unsupervised anomaly detection in industrial datasets. Their methodology enhances detection accuracy and scalability, addressing challenges in analyzing complex, high-dimensional data typical of industrial environments.

Event prediction has been central to industrial maintenance strategies, relying on advanced models to anticipate equipment failures and optimize maintenance schedules. Allen’s temporal logic (\cite{allen1983maintaining}) and Bayesian networks have been widely used to model dependencies between events and quantify uncertainty in predictions. Additionally, frameworks like TEILP (\cite{zhang2023automatic}) utilize temporal knowledge graphs to enhance event reasoning.

The integration of AI for predictive maintenance is further explored in event-driven approaches, such as those leveraging probabilistic models (\cite{petsinis2021analysiskeyflavorseventdriven}). Probabilistic Signal Temporal Logic (PrSTL) extends traditional temporal logic frameworks by incorporating stochastic behaviors into event analysis (\cite{sadigh2016safe}). Similarly, Uncertainty-Aware Signal Temporal Logic (STL-U) accounts for variability in system trajectories, enabling robust fault prediction and proactive interventions in real-world systems (\cite{baharisangari2021uncertaintyawaresignaltemporallogic}). These methodologies emphasize the importance of analyzing event dependencies under uncertainty to enable proactive decision-making.

Frameworks like CodeAct demonstrate how LLMs can autonomously implement predictive models and incorporate uncertainty-aware reasoning, streamlining deployment and enhancing accuracy in industrial scenarios.

\subsection{Conditional-Based Management for Industrial Assets}
AI-driven approaches to predictive maintenance have shown significant promise across industries. For example, Jakubowski et al. (2024) highlight the role of AI in optimizing workflows and preventing equipment failures in the steel industry (\cite{jakubowski2024artificialintelligenceapproachespredictive}). Adaptive conditional-based management systems, such as STL-U (\cite{baharisangari2021uncertaintyawaresignaltemporallogic}), further extend the capabilities of AI by integrating probabilistic reasoning and temporal logic into decision-making processes. STL-U enhances fault detection and maintenance scheduling by accounting for uncertainties in operational data, making it an invaluable tool for managing complex industrial systems.

\section{Business Objects and Analytic Functions}\label{sec:bo_functions}

In this section, we focus on the \textbf{Business Object} and \textbf{Analytic Function} catalogs for industrial asset management, as illustrated in Fig.~\ref{fig:intro-motivation}. Using a specific type of chiller — a closed-loop water-cooled chiller — as a practical example, we demonstrate how to define business objects annotated with natural language (NL) expressions. Additionally, we show how analytical functions operate on these objects to facilitate knowledge extraction, statistical analysis, and decision support, including maintenance predictions and recommendations. The efforts presented in this section lay the foundation for developing the agent-oriented solution.

\begin{figure*}[t!]
    \centering
    \begin{subfigure}{0.25\textwidth} 
        \centering
        \includegraphics[width=\linewidth]{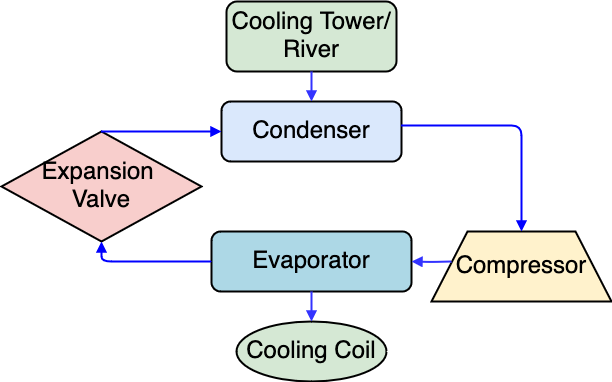}
        \caption{Chiller Diagram showing its three main components of the refrigerant cycle.}
        \label{fig:chiller_draw}
    \end{subfigure}
    \hfill
    \begin{subfigure}{0.35\textwidth} 
        \centering
        \includegraphics[width=\linewidth]{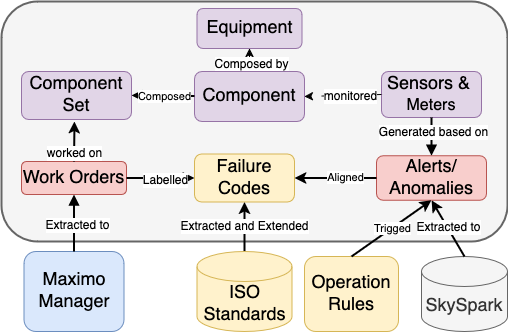}
        \caption{Key Elements of Chiller Asset Management.}
        \label{fig:subfigure2}
    \end{subfigure}
    \hfill
    \begin{subfigure}{0.38\textwidth} 
        \centering
        \includegraphics[width=\linewidth]{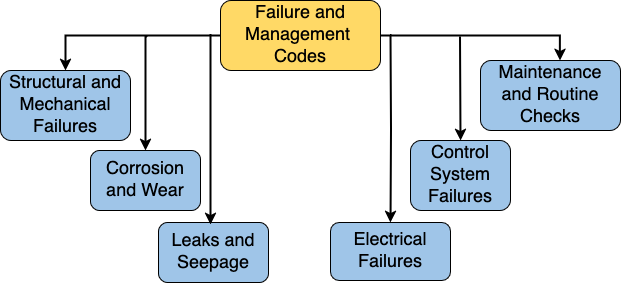}
        \caption{Standardized failure or maintenance Code - Seven Categories and Total 137 Failure and Maintenance Codes.}
        \label{fig:failure_codes}
    \end{subfigure}
    
    \vspace{1em} 
    
    \begin{subfigure}{\textwidth} 
        \centering
        \includegraphics[width=\linewidth]{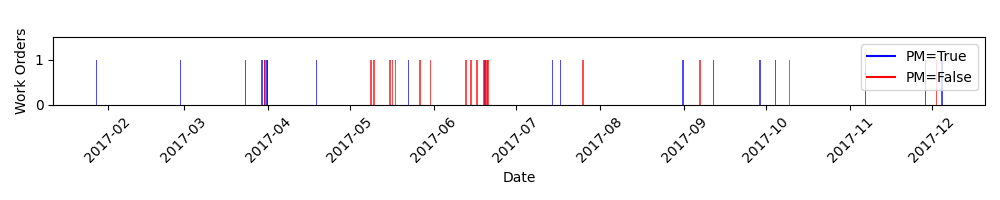}
    \caption{Work Order Events in 2017 for Chiller 9: Preventive Maintenance (Blue - Relative Evenly Distributed) vs. Corrective Maintenance (Red - Clustered in Summer). This pattern suggests periodicity in preventive work orders and seasonal clustering of corrective maintenance tasks.}
    \label{fig:chiller_work_orders_2017}
    \end{subfigure}

    \caption{Asset Management for Chiller}
    \label{fig:data_model}
\end{figure*}

Figure~\ref{fig:data_model}(b) provides a schematic overview of the key concepts. An equipment or asset, such as a chiller (shown in Figure~\ref{fig:chiller_draw}), consists of multiple components and typically is part of a larger system. This equipment is monitored by IoT sensors and evaluated using various metrics (e.g., performance KPIs), which help detect undesired performance or operational issues. For instance, SkySpark~\cite{skyspark2024} is an IoT repository that stores sensor and meter values while generating alerts for predefined conditions. In our study, we utilize real datasets collected from 11 chillers, which serve as the foundation for developing our application.

\subsection{Business Objects Identification and Categorization}

Business objects serve as structured entities that encapsulate domain-specific data, enabling seamless operations and efficient interactions with a Large Language Model (LLM). Defined as Python classes, these objects include detailed descriptions (\textit{Docstring}), maintain standardized attributes, enforce data integrity through validation mechanisms, and provide readable string representations (\texttt{\_\_str\_\_}) to facilitate effective LLM interaction, as LLM interactions are represented as text strings. 

To systematically organize business objects, they are categorized based on their role within the application ecosystem. The primary categories, with examples shown in Table~\ref{tab:business_objects} of Appendix~\ref{appendix:summary_bos_utterances}, include:
\textbf{Content Objects:} Represent fundamental sequences of operational data, such as work orders, alerts, and events over time; \textbf{Meta/Profile Objects:} Store reference information, including asset details, failure modes with corresponding codes, and textual descriptions of failure symptoms; \textbf{Relationship/Causality Objects:} Capture dependencies between events and system behaviors; and \textbf{Recommendation Objects:} Provide suggested work orders with actionable insights and rations to support decision-making (More details can be found at  Appendix~\ref{appendix:summary_bos_utterances}).

\subsubsection{Key Business Objects}

Among all business objects, we focus on three in detail:
\textbf{Failure and Maintenance Codes}, which provide structured classification labels; \textbf{Work Orders}, which encompasses all work orders, and
\textbf{Events}, which contains all recorded events.

\paragraph{Failure and Maintenance Code Objects}:

Currently, many work orders (WOs) in industrial asset management are created manually, including those related to the chillers examined in this research. These WOs lack proper labeling, relying instead on unstructured descriptions that often contain jargon and may be incomplete.  

Leveraging Large Language Models (LLMs) and the ISO standard document~\cite{ISO142242016}, we generate failure and maintenance codes for closed-loop water-cooled chillers. Appendix~\ref{sec:appendix_failure_code_generation} provides more details on creating failure codes. These codes serve as a standardized classification system for unstructured work orders. Since existing WOs are uncategorized, performing advanced analytical analysis without classification is challenging. Standardized failure codes ensure uniformity in maintenance records. Each failure code includes following:
\textbf{Category:} A broad classification of maintenance activities (e.g., Maintenance and Routine Checks);  \textbf{Primary Code:} A high-level failure category within the broader classification (e.g., Major Overhaul);\textbf{Secondary Code:} Specific failure details within the primary category (e.g., Compressor Overhaul); and corresponding \textbf{Description:} A brief explanation of the primary or secondary code.

\paragraph{Work Order Objects}

Work orders document maintenance activities—whether preventive or corrective—while providing essential details such as issue descriptions, associated equipment, priorities, and estimated versus actual labor hours. These objects originate from the work order management system in Maximo Manage~\cite{maximo}. To enable analytical analysis, we developed a method to label previously ``unlabeled'' work orders with failure and maintenance codes (Appendix~\ref{sec:appendix_failure_code_generation}). This classification enhances data structure, enabling more effective insights and decision-making.  

\paragraph{Event Objects}

Event objects consolidate operational occurrences, including maintenance interventions, equipment failures, and system alerts. Unlike isolated work orders or alerts, event objects offer a holistic view of system behavior by aggregating time-series data from multiple sources. These objects serve as a crucial link between individual maintenance tasks and broader operational trends.  

Analyzing event objects allows us to identify recurring patterns, correlate alerts and anomalies with failures, and refine maintenance  strategies. Such event-based insights enhance failure prediction and provide contextual awareness, supporting proactive maintenance planning.

\subsection{Analytic Functions}\label{sec:analytic_functions}

We develop a set of analytic and operational functions (also refer to Appendix~\ref{sec:appendix_analytic_functions}) tailored for work order applications, leveraging diverse business objects. 
Firstly, we created \textbf{Knowledge Extraction Functions}. These functions include retrieving work orders and events, filtering preventive and corrective work orders, and categorizing them based on specified equipment and date ranges. For event extraction, the function retrieves events associated with the selected equipment, event types, and time frame, outputting the results as a string or a pickle file path for the agent to process.

Secondly, we developed \textbf{Statistical Functions}. Those functions in the framework analyze historical maintenance data to extract insights that enhance predictive maintenance, optimize asset reliability, and support data-driven decision-making. \textbf{Work order distribution analysis} quantifies the frequency of failure codes across equipment and time periods, helping prioritize recurring issues and allocate resources efficiently. \textbf{Temporal relationship identification} applies Allen’s Interval Algebra to classify maintenance event sequences (e.g., \textbf{before, after, overlaps}), providing a structured understanding of event dependencies and enabling proactive interventions when alerts consistently precede failures.  Additionally, \textbf{repeated work order pattern detection} identifies cyclic failure intervals, ensuring preventive maintenance schedules align with actual degradation trends rather than fixed time-based schedules. These statistical functions integrate seamlessly into the event-driven framework, allowing AI-driven agents to analyze event sequences, predict upcoming maintenance needs, and refine scheduling strategies. The experiments completed in  Section~\ref{sec:experiment} demonstrates their impact in improving failure forecasting, reducing maintenance costs, and enhancing overall system reliability.

Thirdly, we had \textbf{Decision Support Functions} - (more details available at Appendix ~\ref{sec:appendix_analytic_functions}). These  functions are designed to optimize decision-making by enabling predictive work order generation, alert- and anomaly-based recommendations, and bundled work order identification. 
\textbf{Prediction of Future Work Orders} Utilizes historical maintenance data to forecast the likelihood of various work order types within a specific timeframe, supporting proactive scheduling and resource planning; \textbf{Work Order Recommendations from Alerts} Analyzes rule-based alerts, mapping them to failure categories and prioritizing based on alert frequency and severity to ensure timely preventive maintenance; \textbf{Work Order Recommendations from KPI Anomalies:} Generates corrective maintenance actions by linking key performance indicator anomalies (high or low and frequency) to predefined failure categories and descriptions; and \textbf{Identification of Work Order Bundles:} Suggest grouping related work orders within a set period to consolidate tasks, reduce redundancy, and optimize resource utilization.  The next section discusses integrating the BOs and analytic functions into a \textit{CodeReAct} agent framework. 

\section{Solution Framework - Integration with \textit{CodeReAct} Agent}

Our solution is built upon the \textit{CodeReAct} framework that integrates the Thought-Act-Observation process over a large language model (LLM) agent.  Unlike the typical ReAct agent, our CodeReAct agent functions as a mini-programmer to glue an utterance, BOs, and analytical functions together to complete a task.

\subsection{Framework of CodeReAct Agent}

\begin{figure*}[hbtp]
    \centering
    \begin{subfigure}[t]{0.40\textwidth}
        \centering
        \includegraphics[width=\textwidth]{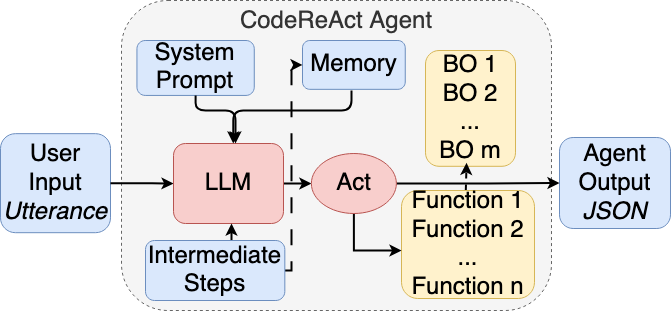}
        \caption{CodeReAct Agent}
        \label{fig:codereact_agent}
    \end{subfigure}
    \hfill
    \begin{subfigure}[t]{0.55\textwidth}
        \centering
        \includegraphics[width=\textwidth]{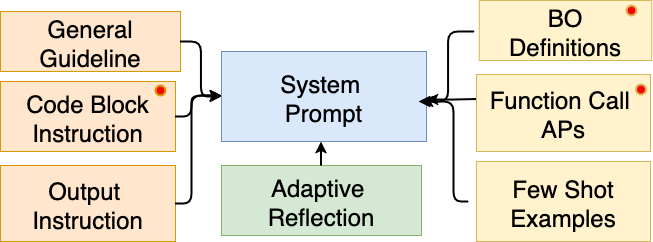}
        \caption{System Prompt (red dot boxes enable code generation)}
        \label{fig:system_prompt}
    \end{subfigure}
    
    \begin{subfigure}[t]{0.80\textwidth}
        \centering
        \includegraphics[width=\textwidth]{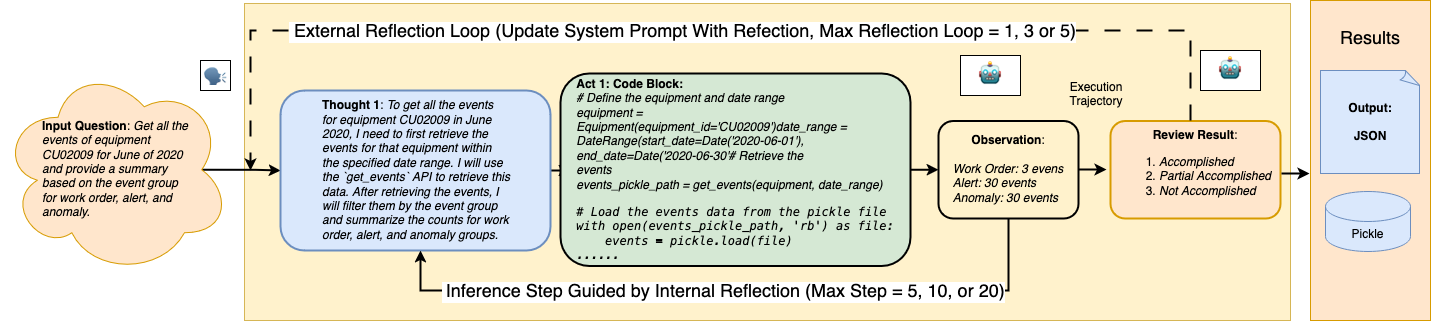}
        \caption{Trajectory of Agentic Process to Solve a Utterance, Such as Chiller Asset Management}
        \label{fig:example_agentic_flow}
    \end{subfigure}    
    \hfill
    \begin{subfigure}[t]{0.80\textwidth}
        \centering
        \includegraphics[width=\textwidth]{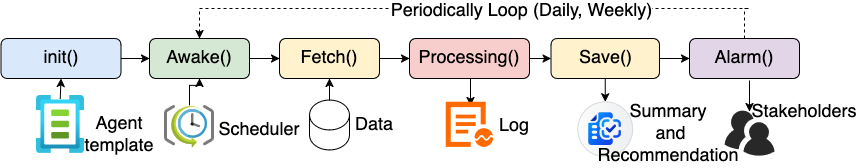}
        \caption{Online Agent and Client Deployment}
        \label{fig:online_agent}
    \end{subfigure}
    
    \caption{Overview of CodeReAct Agent, Agentic Process, and Online Agent Deployment}
    \label{fig:combined_overview}
\end{figure*}

We introduce the key concepts of the framework as follows:\textbf{Utterance:} An utterance is a written query that serves as an input for task requirements (examples are available in Table~\ref{tab:utterance_categorization}). To ensure clarity and efficiency, each utterance must include:  
    1) an asset specification,  
    2) an intended action (e.g., retrieval, summarization, analysis, prediction),  
    3) a defined time frame (required), and  
    4) relevant constraints (optional);  \textbf{System Prompt:} To ensure that CodeReAct agents can correctly execute an utterance, we integrate business objects and analytic function API knowledge into the \textit{system prompt} (as shown in Figure~\ref{fig:system_prompt}). Additionally, the inclusion of a \textit{block Python code} guides the agent's Thought-Action-Observation process, detailing how to:  
    1) analyze information,  
    2) execute actions (via API calls or code), and  
    3) observe results and generate a final response; and \textbf{Agent Process Flow:} Figure~\ref{fig:codereact_agent} illustrates the CodeReAct agent's workflow, highlighting its interactions with user input, system prompts, memory, application functions, and business object (BO) data to produce structured outputs. The workflow integrates intermediate reasoning steps within the LLM and employs a code generation and execution step (``Action'') to dynamically process tasks.  

A natural question arises: what are the advantages of CodeReAct compared to regular ReAct when interacting with pre-defined tools?   As illustrated in Figures~\ref{fig:codereact_agent}, ~\ref{fig:system_prompt}, and~\ref{fig:example_agentic_flow}, CodeReAct first annotate \textit{system prompt} with Python \textbf{code block instruction}. This is followed by a run-time Python code block generation and execution during the utterance processing. Such Python code blocks enhance efficiency, facilitate seamless data exchange, and improve flexibility, making CodeReAct superior to text-based ReAct for multi-step reasoning and analytics.  We summarize the advantages as follows: (1) \textbf{Functions as ReAct Tools}, acting as dynamic, reusable operations; 
(2) \textbf{Seamless Multi-Tool Integration}, combining APIs, business logic, and analytics in a single step; 
(3) \textbf{Self-Sufficient Execution}, enabling code generation and execution on demand; 
(4) \textbf{Reduced LLM Calls}, allowing complex reasoning within one call; 
(5) \textbf{Supports Multi-Entity Operations}, such as batch processing; and 
(6) \textbf{Facilitates Data Integration \& Exchange}, including string-based BOs, Pickle serialization, and Python-based object exchange.

An illustrated example of run-time flow is provided in Figure~\ref{fig:example_agentic_flow} to outline the CodeReAct process, which follows the inner loop of Thought-Action-Observation with outer iteration(s), concluding with a final review. To enhance transparency, we present the detailed algorithm in Algorithm~\ref{alg:appendix_main_algorithm} (detailed in Appendix~\ref{sec:appendix_algorithm}.  To ensure accurate responses to utterances while balancing computational resource requirements, we introduce several innovative enhancements into Algorithm~\ref{alg:appendix_main_algorithm}. These include:  
a) detecting deadlocks in the inner loop,  
b) incorporating a \textit{review} agent to evaluate and categorize outcomes,  
c) dynamically refining the system prompt for subsequent iterations, and  
d) adjusting the LLM model's temperature across iterations to optimize response diversity and stability.  

This flexible approach serves as the foundation for our experiments in the next section.

\subsection{System for Automated Real-Time Work Order Management}

Building on these advantages, we implement an online work order management system to operationalize CodeReAct’s capabilities in real-time asset management. This agent automates work order processing by integrating real-time data analytics and asset management insights (Figure~\ref{fig:online_agent}). Operating within a structured process flow defined by a configuration template, it delivers prioritized tasks, failure insights, and maintenance recommendations. A selected set of utterances ensures meaningful insights, aligning with the needs of the \textit{facility manager}, \textit{reliability engineer}, and \textit{operation engineer}.  The system's workflow consists of the following steps as shown in Figure~\ref{fig:online_agent}.

\section{Experiments}\label{sec:experiment}

This section evaluates the proposed framework through structured experiments, assessing its capabilities based on various combinations of parameters (Table~\ref{tab:model_execution_parameters}), focusing on utterance design and performance analysis.  Additional details of the machine used and utterance execution time can be found at Appendix~\ref{sec:addition_imp_details}.

\subsection{Utterance Selection and Category}

Thirty-six utterances have been carefully crafted to encompass various chiller maintenance tasks, as outlined in Table~\ref{tab:utterance_categorization} (See Appendix~\ref{appendix:summary_bos_utterances}), while incorporating the utilization of BOs and analytical functions discussed in Section~\ref{sec:analytic_functions}. To ensure clarity and efficiency, each utterance must include key elements such as asset specification, intended action (e.g., retrieval, analysis, prediction), a defined time frame, and relevant constraints if required.

Table~\ref{tab:utterance_categorization} categorizes representative natural language utterances related to work order analysis into nine distinct types, including requests to retrieve, summarize, or predict work order data. These categories capture user intents such as getting or reviewing work orders, detecting failures early, analyzing anomalies, and recommending actions. Each category is illustrated with examples and mapped to a set of labeled IDs, providing a foundation for training intent classification or response generation models in industrial asset management scenarios.

\subsection{Experiment Results}

The evaluation consists of three set of experiments. 1) \textbf{Problem-Solving Evaluation}, which assesses the framework's ability to generate appropriate and insightful responses using the utterances in Table~\ref{tab:example_utterances};  2) \textbf{Adaptive Capability Assessment}, which evaluates the framework's capabilities in address an utterance task with the absent of analytic function APIs; and 3)  \textbf{Business Applicability Validation}, which focuses on a real-world chiller maintenance scenario. Table~\ref{tab:model_execution_parameters} outlines the parameters used in these experiments.  

\begin{table}[ht]
\centering
\caption{Experiment Parameters}
\label{tab:model_execution_parameters}
\small
\begin{threeparttable}
\begin{tabular}{|p{1.2 cm}|p{1.0cm}|p{4cm}|}
\hline
\textbf{Parameter} & \textbf{Candidate} & \textbf{Description} \\ \hline
CodeReAct Model & Granite, Mistral-Large & The LLM are used for processing input questions, generating solutions, and executing actions. \\ \hline
Review Model & Granite, Mistral-Large& The LLM is used to review the trajectory of a whole iteration and decide whether the task is completed. \\ \hline
Inner Loop & 10, 20 & Number of steps in the inner loop of iteration \\ \hline
LLM Temperature & 0, 0.1, 0.3, 0.5 & Controls  thought-act-observation process; A higher values expected to break the deadlock of execution. \\ \hline
Outer loop & 1, 3, 5 & Maximum number of reflections to refine execution with update system prompt if iteration > 1. \\ \hline
\end{tabular}
\normalsize
\end{threeparttable}
\end{table}

\subsubsection{Problem-Solving Capability}  

\begin{figure}
    \centering
    \includegraphics[width=0.95\linewidth]{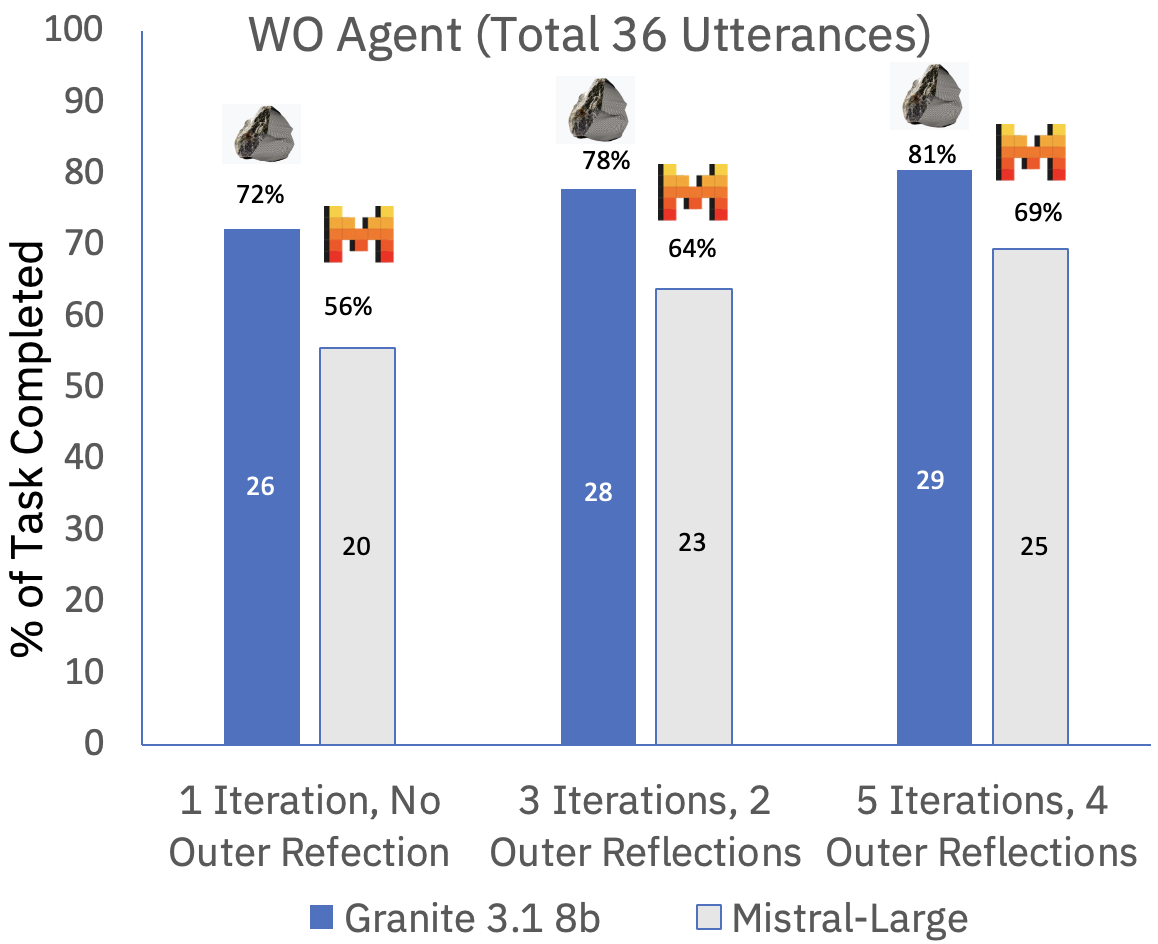}
        \caption{Impact of Outer Reflection (T=0, Inner step = 20)}
        \label{fig:reflection_two_models}
\end{figure}

\begin{figure}[t]
    \centering

    \begin{subfigure}{0.95\linewidth}
        \centering
        \includegraphics[width=\linewidth]{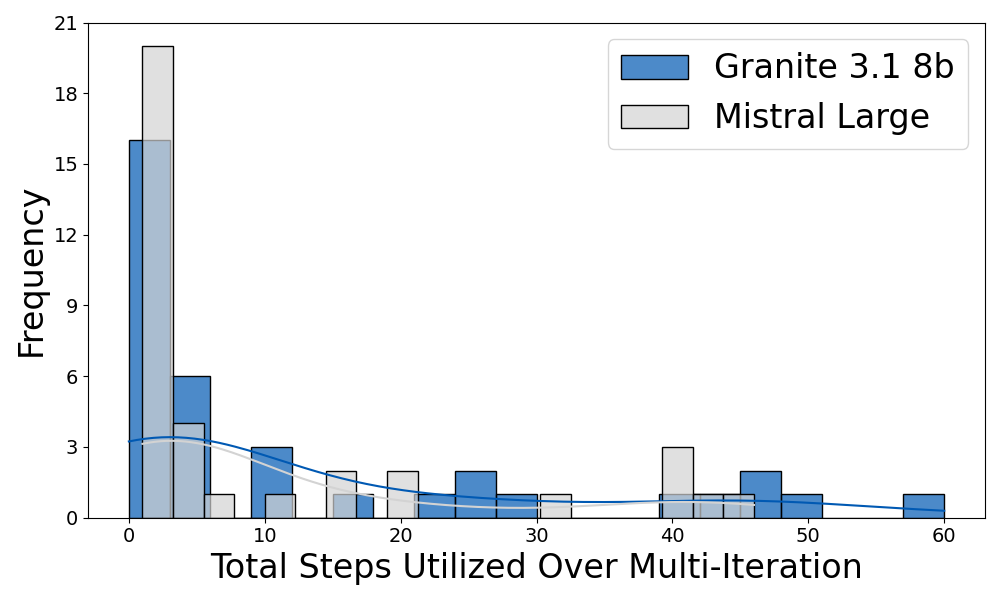}
        \caption{Distribution of \# Steps Utilized for Utterances
        (T=0, Iteration=5)}
        \label{fig:comp_steps_3i}
    \end{subfigure}

    \vspace{0.3cm}

    \begin{subfigure}{0.95\linewidth}
        \centering
        \includegraphics[width=\linewidth]{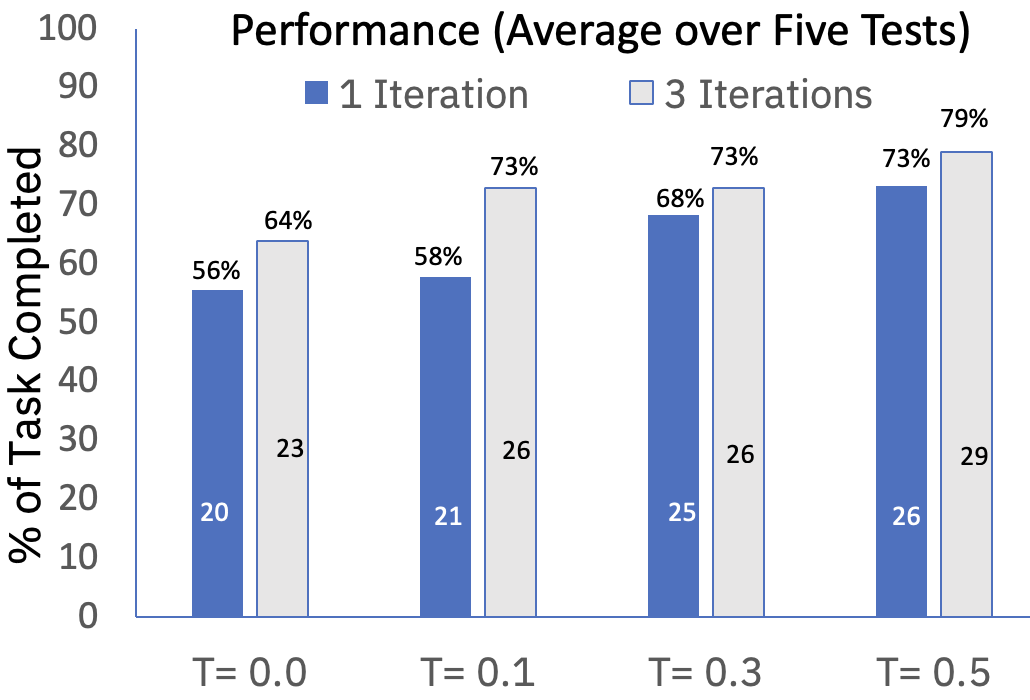}
        \caption{Temperatures/Reflections on Performance
        (Mistral-Large)}
        \label{fig:temperatures_completion}
    \end{subfigure}

    \caption{Performances (T=0, Inner step = 20)}
    \label{fig:combined_figures}
\end{figure}

In this study, we evaluate the framework's ability to complete tasks specified by the thirty-six utterances. First, we employ a \textit{review} agent to assess the response of a task. The review agent classifies the response to \textit{Accomplished}, \textit{Partial Accomplished} and \textit{Not Accomplished}. If the review result is classified as \textit{Accomplished}, we then examine the execution trajectory to verify that the correct functions are invoked with the appropriate parameters. Tasks that do not meet these criteria are listed in Tables~\ref{tab:granite_unaccomplished_tasks} and ~\ref{tab:mistral_unaccomplished_tasks}.  Our experiments show that CodeReAct's performance can be progressively enhanced by leveraging outer iterations, where new reflections are incorporated into subsequent iterations through the updated system prompt. The results are summarized in Figure~\ref{fig:reflection_two_models}. Additionally, the total number of Thought-Act-Observation steps follows a long-tail distribution, as shown in Figure~\ref{fig:comp_steps_3i}, primarily due to deadlocks and difficulties in breaking out of repetitive loops.

\begin{table}
\centering
\caption{Granite-8-1-8b Unaccomplished Tasks (Success rate 72\%,  without Extra Outer Iteration/Reflection)}
\label{tab:granite_unaccomplished_tasks}
\small
\begin{threeparttable}
\begin{tabular}{rll}
\toprule
 ID & - & Review Message \\
\midrule
 1 & P & Returned file path instead of JSON output. \\
 11 & N & Syntax errors prevented work order retrieval. \\
 13 & N & Incorrect date types and insufficient data. \\
 15 & N & Unterminated string error blocked execution. \\
 17 & N & Undefined functions and classes caused failure. \\
 24 & N & Failed to bundle work orders due to missing var. \\
 27 & N & Did not identify alerts due to improper filtering. \\
 29 & N & Invalid code blocked warning message generation. \\
 30 & N & Import error prevented execution of functions. \\
 33 & N & No relevant data found for work order reco. \\
\bottomrule
\end{tabular}
\begin{tablenotes}
\item[*] T=0, Inner step=20, P- Partial accomplishment, N - Not accomplished
\end{tablenotes}
\end{threeparttable}
\normalsize
\end{table}

\begin{table}
\centering
\caption{Mistral-Large  Unaccomplished Tasks (Success rate 56\%, without Extra Outer
Iteration/Reflection)}
\label{tab:mistral_unaccomplished_tasks}
\small
\begin{threeparttable}
\begin{tabular}{p{0.2cm}p{0.1cm}p{6.5cm}}
\toprule
 ID & S &                       Review Message \\
\midrule
  2 &      N & Syntax errors in JSON caused looping. \\
  3 &      N & Datetime handling  led to repetitive failures. \\
  4 &      N & Repeated syntax errors prevented valid JSON. \\
  8 &      P & Looping errors in JSON formatting. \\
  9 &      N & Incomplete code execution with syntax errors. \\
 11 &      N & Incorrect file path returned not the bundled WOs. \\
 12 &      N & Unterminated string literal caused repetitive failures. \\
 13 &      N & JSON formatting errors led to invalid responses. \\
 15 &      N & Looping due to undefined variable 'total\_work\_orders'. \\
 17 &      N & Failure to initialize 'recommended\_primary\_codes' correctly. \\
 18 &      N & Incorrect API handling caused repeated failures. \\
 20 &      N & Looping in JSON formatting. \\
 24 &      N & Failure to execute `find\_bundles` function correctly. \\
 31 &      N & Syntax error in  JSON formatting caused looping. \\
 34 &      P & Failure to check corrective WOs and recommend one. \\
\bottomrule
\end{tabular}
\begin{tablenotes}
\item[*] T=0, Inner step=20, P- Partial accomplishment, N - Not accomplished
\end{tablenotes}
\end{threeparttable}
\normalsize 
\end{table}

To address these challenges, we incrementally increase the model temperature, which improves completion rates (Figure~\ref{fig:temperatures_completion}) alongside additional iterations. These insights enable us to refine the execution algorithm (Algorithm~\ref{alg:appendix_main_algorithm}) by introducing controlled variation in code generation (Action step) through a gradual increase in model temperature as iterations progress.

\subsubsection{Code Adaptive Capability Assessment}

The adaptability relies on the description of the system prompt (details in Figure~\ref{fig:system_prompt}. This experiment introduces a structured methodology to assess the adaptability by updating the system prompt. We selected six tasks with varying levels of complexity, as shown in Table~\ref{tab:code_adaptiability}. The corresponding utterances are presented in Table~\ref{tab:utterance_categorization}. To streamline the evaluation, we removed all analytical function descriptions, retaining only two essential functions for retrieving work orders and events, along with the descriptions of business objects (BOs).

Next, we introduce additional guidelines to two challenging utterances to assess \textbf{CodeReAct}'s capabilities to complete tasks without relying on pre-defined functions. We denote \textbf{U} as the original utterance, while \textbf{U+} and \textbf{U++} represent utterances with progressively detailed instructions (as shown in Table~\ref{tab:utterance_guideline}).   The results, summarized in Table~\ref{tab:code_adaptiability}, indicate that \textbf{CodeReAct}, without assistance from analytical function APIs, successfully handles filtering and summarizing tasks. However, it encounters challenges in solving bundling and work order prediction tasks. By incorporating additional guidelines in the extended utterances, only the bundling task is effectively accomplished.  

\begin{table}[htbp]
    \centering
    \begin{threeparttable}
        \caption{Code Adaptability  Assessment\tnote{0}}
        \label{tab:code_adaptiability}
        \small 
        \begin{tabular}{|p{3.0cm}|>{\centering\arraybackslash}p{.65cm}|>{\centering\arraybackslash}p{.65cm}|>{\centering\arraybackslash}p{.65cm}|>{\centering\arraybackslash}p{.65cm}|}
            \hline
            \textbf{Category} & \textbf{ID} & \textbf{U} & \textbf{U+} & \textbf{U++} \\
            \hline
            Get Work Orders (distribution) & 1 & Yes & - & - \\
            \hline
            Get Work Orders (filtering) & 3 & Yes & - & - \\
            \hline
            Summarize Events & 5 & Yes & - & - \\
            \hline
            Bundle Work Orders\tnote{4} & 24 & Hall & Yes- & Yes \\
            \hline
            Predict Work Orders\tnote{5} & 14 & No & No & Hall \\
            \hline
        \end{tabular}
        \begin{tablenotes}
            \tiny
            \item[0] Mistral Large, T=0.1, inner step=20, no outer reflection, five tests
            \item[1] Yes - Task completed successfully; Yes-: Only one test failed due to Hallucination; Hall - Hallucination output; No - Task not completed 
        \end{tablenotes}
    \end{threeparttable}
\end{table}

\begin{table}[htbp]
    \centering
    \caption{Updated Utterances}
    \label{tab:utterance_guideline}    
    \small
    \begin{tabular}{|p{1.0cm}|p{1.5cm}|p{3.5cm}|}
        \hline
        \textbf{Category} & \textbf{U+} Utterance & \textbf{U++} Utterance (trimmed) \\
        \hline
        Bundle Work Orders (ID 24) & (\textbf{U}) + Bundles should not have overlap. & (\textbf{U}, \textbf{U+}) + Group corrective work orders based on actual finish dates within an allowed time gap. Discard bundles with only one work order. \\
        \hline
        Predict Work Orders (ID 14) & (\textbf{U}) + Using a Markov approach and specify the state with the primary failure code. & (\textbf{U}, \textbf{U+}) + Use a Markov model to forecast work orders based on state transitions from historical data and update predictions using transition probabilities. \\
        \hline
    \end{tabular}
    \normalsize    
\end{table}

\subsubsection{Ablation Study: Impact of the \textit{Act} Only without \textit{Thought}}

To evaluate the role of reasoning in CodeReAct, we conducted an ablation by removing the \textit{Thought} step, leaving only \textit{Act--Observe} (akin to CodeAct). Using the Granite-3.1-8b model and 36 utterances, we observed a sharp drop in task completion when reasoning was removed (Table~\ref{tab:ablation}). Each configuration was tested with a fixed temperature of 0.0, 20 inner-loop steps, and varying outer-loop iterations (\{1, 3, 5\}).
This highlights that executable code alone is insufficient; intermediate reasoning is essential for grounding actions and improving reliability.

\begin{table}[ht]
\centering
\caption{Task completion with and without \textit{Thought} step }
\label{tab:ablation}
\small
\begin{tabular}{lccc}
\toprule
\textbf{Outer Iters} & \textbf{Full} & \textbf{Act-Only} & \textbf{Drop} \\
\midrule
1 & 24 & 12 & -50\% \\
3 & 28 & 16 & -43\% \\
5 & 29 & 20 & -31\% \\
\bottomrule
\end{tabular}
\end{table}

\section{Conclusion}\label{sec:colclusion}

In this paper, we present CodeReAct, an LLM-driven agentic framework that automates industrial work order analytics through executable code generation and natural language interaction. The system reveals actionable maintenance patterns—such as seasonal trends, key alert impacts, and bundling opportunities—demonstrated through a chiller case study (Appendix~\ref{sec:appendix_client_insight_experiment}). While evaluated on chillers, the framework generalizes to AHUs, pumps, generators, and other industrial assets, illustrating its cross-domain applicability. By recontextualizing NLP agent frameworks for operational decision-making, CodeReAct shows how NLP advances can extend beyond traditional language tasks. Future work includes refining the business object and analytic function framework and conducting self-consistency testing to improve robustness across varied utterances and real-world settings.

\bibliography{custom}


\clearpage
\appendix

\section{Failure Code Generation and Work Order Normalization Categorization}\label{sec:appendix_failure_code_generation}

\begin{table}[ht]
    \centering
    \caption{Work Order Data Categories}
    \label{tab:work_order_data}
\renewcommand{\arraystretch}{1.3}
    \begin{tabular}{|p{1.8cm}|p{5.3cm}|}
        \hline
        \textbf{Category} & \textbf{Details} \\
        \hline
        \textbf{Asset Information} & Asset ID (equipment identifier), Location, and Site \\
        \hline
        \textbf{Work Details} & Work Description (task explanation), Work Type (e.g., Emergency, Preventive, Predictive) \\
        \hline
        \textbf{Priorities} & Work Priority (urgency of the task), Asset Priority (importance of the equipment) \\
        \hline
        \textbf{Scheduling} & Reported Date, Target Start/Finish, Actual Finish, Duration, Actual Labor Hours \\
        \hline
        \textbf{Work Status} & Status of the task (e.g., Pending, Ready to Start, Completed) \\
        \hline
        \textbf{Linked Data} & Connects to asset details for consistency (e.g., location, priority, attributes) \\
        \hline
    \end{tabular}
\end{table}

This effort includes three steps.  The first step is to prepare a consistent three-level hierarchical failure or maintenance code. This is followed by automatic work order labeling using the work order description and the failure or maintenance codes.  Finally, based on the alert description, we associate each alert with the failure or maintenance codes.

\begin{figure*}[htbp]
    \centering
    \includegraphics[width=0.80\textwidth]{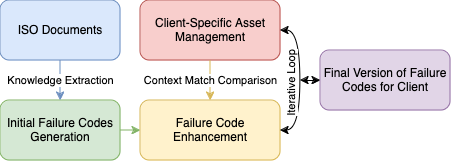}
    \caption{failure or maintenance Code Generation with LLM}
    \label{fig:failure_code_generation}
\end{figure*}

As shown in Figure~\ref{fig:failure_code_generation}, the failure or maintenance code generation process begins with Large Language Models (LLMs) extracting knowledge from ISO 14224-2016 \cite{ISO142242016}. This knowledge is used to generate an initial set of failure or maintenance codes, organized into three level hierarchy (see Figure~\ref{fig:failure_codes}). 

The failure or maintenance codes generated using ISO document may not fully cover all client scenarios. To address this, the work order descriptions with low confidence in failure or maintenance code assignment is analyzed. The low-confidence cases prompt the generation of additional failure or maintenance codes, which are incorporated and refined through context-match comparisons. 

To address this, we leverage large language model (LLM) to suggest potential failure or maintenance codes based on the descriptions of the rules triggering the alerts. The LLM analyzes the rule descriptions and context within the alerts, aiding in establishing a more precise link between the alert signals and the likely failure or maintenance code(s). For example, an alert indicating that a chiller's load is too low may correspond to multiple potential failure or maintenance codes related to performance issues.

\begin{table}[ht!]
\caption{Examples of Primary \& Secondary Codes (Category: Maintenance and Routine Checks) }\label{tab:appendix_sample_failure_codes}
\centering
\small 
\begin{tabular}{|p{0.15\textwidth}|p{0.21\textwidth}|}
\hline
\textbf{Code} & \textbf{Description} \\
\hline
MT006 & Major Overhaul \\
\hspace{0.5em}MT006a & \hspace{0.5em}Compressor Overhaul \\
\hline
MT013 & Vibration Analysis \\
\hspace{0.5em}MT013b & \hspace{0.5em}Routine Vibration Analysis \\
\hline
MT014 & Filter Replacement \\
\hspace{0.5em}MT014c & \hspace{0.5em}Water Filter Replacement \\
\hline
\end{tabular}
\end{table}

\section{CodeReAct Algorithm with Outer Loop of Iteration for Refinement}\label{sec:appendix_algorithm}

The CodeReAct Algorithm (Algorithm~\ref{alg:appendix_main_algorithm}) is designed to process natural language queries (utterances) by iteratively generating code, executing it, and refining the approach based on observations. A key feature of this algorithm is its ability to execute generated Python code, such as data processing and function API calls, during the Action Execution phase. The algorithm breaks the inner loop if a deadlock is detected in the thought-action-observation cycle. It proceeds to the review phase, updating the system prompt with reflections for the next iteration.

\begin{algorithm}[H]
\caption{CodeReAct Algorithm with Outer Loop of Iteration for Refinement}\label{alg:appendix_main_algorithm}
\begin{algorithmic}[1]
\STATE \textbf{Input:} \texttt{query} (Natural Language Utterance)
\STATE \textbf{Parameters:} LLM models, max\_steps, max\_reflection\_loops, list of temperatures (optional)
\STATE \textbf{Output:} Final result (JSON or Pickle format)
\STATE Initialize the system prompt with \texttt{query}

\FOR{\texttt{reflection\_loop = 1 to max\_reflection\_loops}}
    \FOR{\texttt{step = 1 to max\_steps}}
        \STATE \textbf{Thought Generation:}
        \STATE \quad Generate a structured thought based on the query and previous thought-act-observation step
        \STATE \quad Plan \textbf{coding} actions needed to complete the task

        \STATE \textbf{Action Execution:}
        \STATE \quad {\color{red}{Execute the generated Python code (e.g., data processing), function API calls}}

        \STATE \textbf{Observation:}
        \STATE \quad Analyze the output of the executed action
        \STATE \quad Record observations (e.g., counts of work orders, alerts, anomalies, suggestions)
        \IF{\texttt{deadlock detected}}
            \STATE {\color{red}{Break the inner loop and proceed to the review phase}}
        \ENDIF
    \ENDFOR

    \STATE \textbf{Review} the entire trajectory of this iteration, and \textbf{classify} results as: Accomplished, Partially Accomplished or Not Accomplished
    \IF{\texttt{result == Accomplished}}
        \STATE Break the outer loop and proceed to finalize output
    \ELSE
        \STATE \quad {\color{red}{Update the system prompt with reflection}}
    \ENDIF
\ENDFOR

\STATE Generate final output, save as JSON with optional Pickle file
\STATE \textbf{Return:} Final result
\STATE Log the entire execution trajectory from the last iteration.

\end{algorithmic}
\end{algorithm}

\section{Prompt Design}

To enable the CodeReAct agent to generate executable code in response to natural language queries, we adopt a few-shot prompting strategy. These few-shot examples serve as scaffolding, demonstrating how to decompose a user query into structured reasoning and corresponding code. This approach is crucial for grounding the agent’s behavior, ensuring that it produces valid, context-aware actions when solving domain-specific analytical tasks.

Each few-shot example is structured around the \textit{Thought-Action-Observation} (TAO) framework, where the agent:
\begin{itemize}
    \item Interprets a natural language question.
    \item Generates intermediate \textbf{Thoughts} to plan the task.
    \item Produces executable Python \textbf{Actions}.
    \item Observes execution outputs or errors (\textbf{Observation}).
    \item Concludes with a structured \textbf{Final Answer}.
\end{itemize}

These few-shot trajectories are included in the system prompt to condition the language model and guide its reasoning and code generation. The following example illustrates how the agent processes a work order retrieval request:

\begin{lstlisting}[caption={Few-Shot Prompt Example: Preventive Work Order Retrieval}]
Question: I would like to retrieve the preventive work order details 
for the equipment labeled as CU02013 for the year 2017. 
Store the result in a file and provide a pointer to the file.

Thought 1: Retrieve all work orders using `get_work_orders`, 
then filter for preventive ones with `filter_preventive_work_orders`.

Action 1:
equipment = Equipment(equipment_id="CU02013")
date_range = DateRange(start_date=Date("2017-01-01"), end_date=Date("2017-12-31"))

work_orders_pickle_path = get_work_orders(equipment, date_range)
with open(work_orders_pickle_path, "rb") as file:
    work_orders = pickle.load(file)

preventive_work_orders = filter_preventive_work_orders(work_orders)

Thought 2: Store the filtered results in a JSON file.

Action 2:
file_path = "preventive_work_orders_CU02013_2017.json"
preventive_work_orders_dict = [vars(pwo) for pwo in preventive_work_orders]

# Handle datetime serialization
for pwo in preventive_work_orders_dict:
    pwo['actual_finish'] = pwo['actual_finish'].strftime('%Y-%m-%d %H:%M:%S')

with open(file_path, "w") as file:
    json.dump(preventive_work_orders_dict, file, indent=4)

Thought 3: Return the file path as a structured final answer.

Action 3:
Final Answer:
{
    "file_path": "preventive_work_orders_CU02013_2017.json"
}
\end{lstlisting}

This example demonstrates how few-shot prompts enable the agent to reason through multi-step queries, recover from execution failures (e.g., timestamp serialization), and produce valid structured outputs.

In addition, the system prompt provides the agent with contextual information, tool descriptions, iteration strategy, and formatting requirements. The agent is positioned as an expert analyst with access to APIs that retrieve and analyze industrial maintenance data.  We define a standalone Python file that includes the domain-specific business object definitions (e.g., \texttt{WorkOrder}, \texttt{Event}, and \texttt{FailureCode} \texttt{DateRange}) and API function signatures used in the work order and alert analysis tasks. This file intentionally omits execution logic, serving instead as an interface specification that informs the language model of the available data types and callable operations. The file is embedded directly into the system prompt, enabling the agent to generate valid, type-aware Python code in alignment with the system’s analytical APIs. The detailed discussion of the Businsss objects and the functions can be found at Section~\ref{sec:bo_functions} and Appendix~\ref{appendix:summary_bos_utterances}.

\section{Summary and Categorization of Business Objects and Utterances}{\label{appendix:summary_bos_utterances}}

This appendix provides additional context on the business object taxonomy and utterance patterns used in the CodeReAct framework.

Business objects, summarized in Table~\ref{tab:business_objects}, form a structured taxonomy for industrial asset management. They are categorized into four types: Content Objects, Meta/Profile Objects, Relationship/Causality Objects, and Recommendation Objects. These objects represent key entities such as work orders, alerts, equipment metadata, and failure codes, each playing a distinct role in data collection, standardization, event linking, and decision support.

Utterances, shown in Table~\ref{tab:utterance_categorization}, capture common natural language queries aimed at gaining insights into asset behavior. These include tasks such as reviewing past issues, detecting early failures, and recommending maintenance actions. Each utterance is tied to a specific analytic intent and typically involves the functional use of one or more business objects for reasoning, retrieval, or forecasting.

Together, the taxonomy of business objects and categorization of utterances define a unified framework for mapping user intents to structured analytic procedures in the CodeReAct system.

\begin{table*}[!ht]
\captionsetup{skip=2pt}
\caption{Summary of Business Objects, Source, Role, and Number of Records}
\label{tab:business_objects}

\begin{threeparttable}
\centering

\begin{tabular}{@{} p{2.8cm} p{2.8cm} p{8cm} c @{}}
\toprule
\textbf{Business Object} & \textbf{Source} & \centering \textbf{Role} & \textbf{Count} \\
\midrule
\multicolumn{4}{@{}l}{\cellcolor{gray!20} \textbf{Content Objects}} \\
WorkOrder & Maximo Manage & Tracks scheduled and unscheduled maintenance tasks, categorized as preventive or corrective. & 4392 \\
Alert Events & SkySpark & Logs real-time alerts triggered by IoT sensors based on predefined conditions. & 1995 \\
Anomaly Events & ML Generated & Detects KPI deviations using machine learning for predictive maintenance. & 542 \\
Event & Aggregated by Authors & Consolidates event logs for tracking and decision-making. & 6929 \\
\midrule
\multicolumn{4}{@{}l}{\cellcolor{gray!20} \textbf{Meta/Profile Objects}} \\
ISO\_Failure\_Code & Developed by Authors & Standardizes failure classification for structured maintenance analysis. & 137 \\
ISO\_Primary Failure\_Code & Developed by Authors & Defines primary failure categories and links related secondary codes. & 68 \\
AlertRule & SME Provided & Specifies conditions for triggering alerts based on system behaviors. & 77 \\
Equipment & SME Provided & Represents industrial assets, including status and specifications. & 22 \\
\midrule
\multicolumn{4}{@{}l}{\cellcolor{gray!20} \textbf{Relationship/Causality Objects}} \\
Alert-Rule Mapping & Relationship & Links alert rules to failure codes for automated diagnostics. & 46 \\
Anomaly Mapping & Relationship & Associates anomalies with failure-codes for predictive insights & 12 \\
\midrule
\multicolumn{4}{@{}l}{\cellcolor{gray!20} \textbf{Recommendation Objects for Work Order Assignment}} \\
Recommendation & Recommendation & Suggests maintenance actions based on historical patterns. & N/A \\
\bottomrule
\end{tabular}

\begin{tablenotes}
    \item[*] The design and the structure of the business objects and corresponding analysis in this section are valid for other industrial asset types, such as standby generators.
\end{tablenotes}
\end{threeparttable}
\end{table*}

\begin{table*}[ht]
\caption{Categorization and Examples of Utterances}
\label{tab:utterance_categorization}
\centering
\small
\begin{tabular}{p{2.5cm}p{10cm}p{2.5cm}}
\toprule
\textbf{Utterance Category} & \textbf{Example} & \textbf{Count [IDs]} \\
\midrule
Get Work Orders & "I would like to check the work order distribution for the equipment labeled as CU02013 for the year 2017 based on primary code." (ID: 1) \newline "I would like to retrieve the corrective work order details for the equipment labeled as CU02013 for the year 2017." (ID: 3) & 4 [0, 1, 2, 3] \\
Summarize Events & "Provide a summary of all events for equipment CU02009 in June 2020, categorized by work order, alert, and anomaly." (ID: 5) & 7 [4, 5, 6, 7, 8, 9, 10] \\
Find Important Alerts & "Given that 80\% of alerts are likely spurious, can you identify the meaningful 20\% for further analysis for any chillers?" (ID: 27) & 1 [27] \\
Review Past Issues & "In May and June 2020, numerous anomalies and alerts were reported for Chiller 9 (equipment ID CU02009). Could you review them and suggest appropriate work orders or actions?" (ID: 21) & 4 [20, 21, 22, 23] \\
Detect Failures Early & "We are developing an early detection system for potential condenser water-side fouling in Chiller 9. Monitor Chiller 9 at the POKMAIN site from 2020-04-20T12:30:00 to 2020-05-12T19:30:00, identify anomalous behaviors, and provide a work order recommendation if no corrective work order has already been created." (ID: 34) & 3 [30, 33, 34] \\
Predict Work Orders & "Assuming the current date is 2018-01-02, can you predict the next work order probability for equipment CU02009?" (ID: 14) & 3 [12, 13, 14] \\
Recommend Work Orders & "An anomaly, 'KPI Cooling Load above upper bound,' has been detected for equipment CU02009. Can you recommend the top three work orders to address this issue?" (ID: 17) & 5 [15, 16, 17, 18, 19] \\
Bundle Work Orders & "Can you provide guidance on bundling corrective work orders to optimize the maintenance schedule for Chiller 9 (equipment ID CU02009) for the years 2017, 2018, and 2019? Each bundle should occur within two weeks, include at least two work orders, and detail each bundle's specifics." (ID: 24) & 2 [11, 24] \\
Analyze Anomalies and Alerts & "How can I analyze anomalies across multiple KPIs to better diagnose the root cause of these issues?" (ID: 25) & 5 [25, 26, 28, 29, 35] \\
\bottomrule
\end{tabular}
\normalsize
\end{table*}

\section{Event Representation and Analytic Functionalities}\label{sec:appendix_analytic_functions}

Machine learning and deep learning algorithms are widely used to predict and forecast water-cooled chiller power consumption, with ML offering simplicity and interpretability, while DL provides higher accuracy for complex, time-series data. Comparative studies highlight the trade-offs between these approaches and emphasize the importance of data quality, model design, and real-time applicability~\cite{chaerun2021comparative}.    
We now define two building blocks for implementing LLM agentic workflow: a) Unified Events and b: Tasks (or Tools). Events represent the core data aspect, whereas tasks enable the extraction of descriptive, predictive, and relational insights from the events.

\subsection{Unified Events Representation}\label{sec:event_representation}
We combined work order and alert events to create a richer dataset. Mathematically, we can define the unified event as follows: 

\begin{equation}
e = (i, t, d, a, g, k)
\end{equation}

\begin{itemize}
    \item $i$:  Event ID
    \item $t$: Start timestamp of the event
    \item $d$: Duration of the event
    \item $a$: Asset, equipment or component
    \item $g$: Event group based on \textbf{P}reventive or \textbf{C}orrective Action or \textbf{Al}lert, and \textbf{An}omaly, (value of {P, C, A1, An})
    \item $k$: Type of events (Failure/Maintenance code for repair or maintenance work order, alert rule for alerts)
    
\end{itemize}


This enriched dataset allows for comprehensive analysis and filtering, enabling the identification of patterns and relationships between different events. For example, it can reveal which alerts ($g = ALERT$) are most predictive of certain work orders ($g = WORK\_ORDER$), or the time intervals between recurring alerts and subsequent maintenance activities. Such insights are crucial for identifying the root causes of failures and enhancing the system's ability to predict future maintenance needs. 


\subsection{Event Pair Relationship Formulation}



By leveraging Allen’s Interval Algebra, we can deepen our understanding of event sequences, anticipate potential issues, and optimize the scheduling of maintenance activities. To better reflect practical scenarios, we will refine the logic of Allen's Interval Algebra. This refinement ensures that relationships like ``before'' and ``after'' are evaluated within a time range, say $\Delta$ a month, that accurately mirrors the physical reality, thereby confirming that the events are genuinely related and relevant to the maintenance context.
Let \( E = \{e_1, e_2, \dots, e_n\} \) be the set of all events in a given context. For any two events \( e_1 = (i_1, t_1, d_1, a_1, g_1, k_1)\) and \( e_2 = (i_2, t_2, d_2, a_2, g_2, k_2)\) from \( E \), their temporal relationship can be described by Allen's Interval Algebra. The three interesting relationships in our problem setting are:

\begin{itemize}
    \item \textbf{Before}: \( e_1 \) ends before \( e_2 \) starts.
    \[
    e_1 \text{ before } e_2 \iff {(t_1+d_1)} < {t_2} \text{ and } {(t_2 - t_1)} \leq \Delta
    \]
    \item \textbf{Overlaps}: \( e_1 \) starts before \( e_2 \), and \( e_2 \) starts before \( e_1 \) ends.
    \[
    e_1 \text{ overlaps } e_2 \iff {t_1} \leq t_2 \leq {t_1+d_1} 
    \]
    \item \textbf{During}: \( e_1 \) starts after \( e_2 \) starts and ends before \( e_2 \) ends.
    \[
    e_1 \text{ during } e_2 \iff {t_2} \leq {t_1} \text{ and } {t_1+d_1} \leq {t_2 + d_2}
    \]
\end{itemize}

In practice, this mathematical framework is applied to analyze the temporal sequence of maintenance work orders and sensor-generated alerts. For instance:

\begin{itemize}
    \item \textbf{Before}: If an alert (e.g., temperature exceeding a threshold) occurs before a maintenance work order, this may indicate a predictable sequence where the alert triggers maintenance.
    
    \item \textbf{Overlaps}: If two work orders overlap, it may indicate simultaneous maintenance activities, which could require coordinated resource allocation.
    
    \item \textbf{During}: An alert occurring during a work order may suggest the need for immediate attention or adjustment during maintenance activities.
\end{itemize}



\subsection{Predicting the Next Work Order with Markov Approach}

A key concern in asset management is anticipating the next necessary work orders. Various algorithms can be employed for this prediction. This task uses a Markov approach to leverage historical work orders to forecast the likelihood of upcoming work order types.

\begin{itemize}
    \item \textbf{State Space:} Let \( S = \{s_1, s_2, \dots, s_n\} \) be the set of all possible work order types.
    \item \textbf{Transition Probability Matrix:} Define the transition probability matrix \( P \), where each element \( P_{ij} \) is given by:
\begin{equation}
    P_{ij} = \Pr(X_{t+1} = s_j \mid X_t = s_i)  
\end{equation}
    \item \textbf{Initial State:} The initial state distribution \( \pi_0 \) is defined as:
    \[
    \pi_0(s_i) = \Pr(X_0 = s_i)
    \]
    \item \textbf{Prediction of Next Work Order Type:} Given the current work order type \( X_t = s_i \), the predicted next work order \( \hat{X}_{t+1} \) is:
    \begin{equation}       
    \hat{X}_{t+1} = \arg\max_{s_j \in S} P_{ij}
    \end{equation}
    \item \textbf{Recurrence Relation:} The probability distribution over the next work order \( X_{t+1} \) can be computed as:
    \begin{equation}
    \pi_{t+1} = \pi_t \cdot P        
    \end{equation}
\end{itemize}

\subsection{Bundle Identification}

We define the bundle identification process mathematically to group work orders based on their actual finish dates. The selected work orders (WO) are \textit{corrective} ones.

\subsubsection{Input Definitions}
Let:
\begin{itemize}
    \item \( W = \{ w_1, w_2, \dots, w_n \} \) be the set of \textit{corrective} work orders, where each work order \( w_i \) has an associated \textit{completion date} \( T_i \).
    \item \( T_i \) represents the \textit{actual finish date} of work order \( w_i \).
    \item \( \Delta \) represents the \textit{maximum allowable interval} (in days) between consecutive work orders for them to be grouped in the same bundle.
    \item \( B_k \) represents the \textit{\( k \)-th bundle}.
\end{itemize}

\subsubsection{Sorting Step}
The work orders are sorted in ascending order based on their \textit{actual finish dates}:
\begin{equation}
    T_1 \leq T_2 \leq \dots \leq T_n.
\end{equation}

\subsubsection{Bundling Process}
The bundling process follows these steps:

\begin{itemize}
    \item Initialize the first bundle:
    \begin{equation}
        B_1 = \{ w_1 \}, \quad b_1 = 1,
    \end{equation}
    where \( b_1 \) is the \textit{bundle identifier}.

    \item For each work order \( w_i \) with \( i > 1 \), compute the time difference between its finish time and that of the previous work order:
    \begin{equation}
        \delta_i = T_i - T_{i-1}.
    \end{equation}

    \item If \( \delta_i \leq \Delta \), add \( w_i \) to the \textit{current bundle}:
    \begin{equation}
        B_k = B_k \cup \{ w_i \}.
    \end{equation}

    \item If \( \delta_i > \Delta \), start a \textit{new bundle}:
    \begin{equation}
        B_{k+1} = \{ w_i \}, \quad b_{k+1} = b_k + 1.
    \end{equation}
\end{itemize}

\subsubsection{Filtering Step (Removing Single-Element Bundles)}
After forming all bundles, remove any bundle \( B_k \) where:
\begin{equation}
    |B_k| = 1.
\end{equation}

\subsubsection{Final Output}
The algorithm returns a set of bundles:
\begin{equation}
    B = \{ B_1, B_2, \dots, B_m \},
\end{equation}
where each bundle \( B_k \) contains at least two work orders grouped according to the interval constraint \( \Delta \), ensuring that \textit{single-element bundles} are excluded.

\section{Insight for the Chill Application}\label{sec:appendix_client_insight_experiment}

The utterance or utternance structure is designed to assess the capabilities of language models in forecasting maintenance events and analyzing work order distributions. This approach simulates real-world scenarios where predictions or insights are crucial for effective asset management. A typical utternance is structured as follows:

\begin{enumerate}

    \item \textbf{Context Definition:} Establish the operational context, such as the current status and maintenance history of a specific asset. For example, ``Chiller 13 has undergone routine maintenance and a compressor overhaul in the past six months.''
    \item \textbf{Event Specification:} Specify the maintenance event or sequence of events to be predicted or analyzed, such as work orders or alerts. For instance, ``Predict the next work order type based on recent maintenance activities and sensor alerts.''
    \item \textbf{Temporal Frame:} Define the period of interest, such as a retrospective analysis of the past year or a forecast for the next quarter. For example, ``What is the likely next work order type within the next two months?''
    \item \textbf{Prediction or Insight Request:} Request a prediction or analysis, asking the model to forecast the next maintenance action or analyze the distribution of past work orders to identify patterns or inefficiencies. For instance, ``Predict the next work order type for Chiller 13.''
    \item \textbf{Expected Output Format:} Specify the expected format of the output, such as a probability distribution, summary report, or list of maintenance recommendations. For example, ``Provide a ranked list of likely work order types with associated probabilities.''
\end{enumerate}

\begin{table}[htbp]
\caption{Example Utterances}
\label{tab:example_utterances}

\centering
\scriptsize 
\arrayrulecolor{black} 
\renewcommand{\arraystretch}{1.1} 
\setlength{\tabcolsep}{4pt} 

\begin{tabularx}{\columnwidth}{|p{0.23\columnwidth}|p{0.71\columnwidth}|} 
\hline
\rowcolor{gray!30}
\textbf{Utternance} & \textbf{Context and Additional Information} \\ \hline

\textbf{Utternance 1:} \textit{``Given the recent maintenance history of Chiller 13 and the last two years of work order data, what is the most likely next work order type within the next two months?''} & 

\setlength{\parskip}{0pt} 
\setlength{\itemsep}{0pt} 
\setlength{\topsep}{0pt} 
\vspace{-1em} 
\begin{itemize}
    \item \textbf{Context Definition:} Recent maintenance actions on Chiller 13, including routine checks and specific repairs.
    \item \textbf{Event Specification:} Focuses on predicting the next work order type based on historical data.
    \item \textbf{Temporal Frame:} Two-month forecast period.
    \item \textbf{Prediction Request:} Most likely next work order type.
    \item \textbf{Expected Output Format:} A ranked list of work order types with probabilities.
\end{itemize} \\ \hline

\textbf{Utternance 2:} \textit{``What has been the distribution of work event types for Chiller 13 over the past year, and how might this inform the scheduling of future maintenance?''} & 
\setlength{\parskip}{0pt} 
\setlength{\itemsep}{0pt} 
\setlength{\topsep}{0pt} 
\vspace{-1em} 
\begin{itemize}
    \item \textbf{Context Definition:} Operational context of Chiller 13 over the past year, including all recorded WOs.
    \item \textbf{Event Specification:} Analysis of the distribution of WO event types (e.g., routine maintenance, emergency repairs, system alerts).
    \item \textbf{Temporal Frame:} The past year.
    \item \textbf{Insight Request:} Analysis to inform future maintenance scheduling.
    \item \textbf{Expected Output Format:} Summary report or visualization showing the frequency and types of WO events.
\end{itemize} \\ \hline

\end{tabularx}
\end{table}

Similarly, the following utterances/results can be framed to leverage the capabilities of tools specified in Table~\ref{tab:example_utterances}. 

\begin{itemize}

    \item \textbf{Allen Relationship Identification:} 
    \begin{itemize}
        \item \textbf{Utternance:}``\textit{What are the temporal relationships between recent work orders and alerts for Chiller 9?}''
        \item \textbf{Result:} The distributions of WOs (aggregated over alert types). Table~\ref{tab:impact_alert_wos} shows the results for Chiller 9.
    \end{itemize}
    
    \item \textbf{Work Order Bundling:} 
    \begin{itemize}
        \item \textbf{Utternance:} ``\textit{Which \textbf{corrective} work orders for Chiller 9 can be bundled in the next maintenance window?}''
        \item \textbf{Result:} For Chiller 9 and 13, we found 5 and 4 bundles respectively with bundle size from 2 to 6. Within a bundle, the two consecutive WOs have their interval less than a week.
    \end{itemize}
    
    \item \textbf{Identify Repeated Patterns for Work Orders:} 
    \begin{itemize}
        \item \textbf{Utternance:} ``\textit{Are there any repeated maintenance intervals for Chiller 13 that suggest recurring issues?}''
        \item \textbf{Result:} No obvious repeated pattern found for Chiller 9 and 13 for any failure/maintenance code.
    \end{itemize}

\end{itemize}

\begin{table}[ht]

\caption{Impact of Alerts over  WOs for Chiller 9 in Two Month Windows (Chiller 9 has 4 year of operation data)}
\label{tab:impact_alert_wos}

\centering
\scriptsize
\begin{tabular}{|p{0.10\textwidth}|p{0.32\textwidth}|}
\hline
\textbf{Alert Type (Occurrences)} & \textbf{Work Orders (Occurrences and Percentage)} \\ \hline
CR00007 - Load Low (11) & 
MT011: Calibration (12, 31.58\%), 
MT010: Oil Analysis (11 occurrences, 28.95\%), 
MT013: Vibration Analysis (11, 28.95\%), 
MT004: Refrigerant Addition (4, 10.53\%) \\ \hline

CR00008 - Evaporator Approach High (183) & 
MT013: Vibration Analysis (301, 37.30\%), 
MT011: Calibration (213, 26.39\%), 
MT010: Oil Analysis (110, 13.63\%), 
MT004: Refrigerant Addition (105, 13.01\%), 
MT001: Routine Maintenance (64, 7.93\%), 
MT009: Refrigerant Transfer (7, 0.87\%), 
MT003: Lubrication (5, 0.62\%), 
E006: VFD (Variable Frequency Drive) Failure (2, 0.25\%) \\ \hline

CR00012 - Chiller Cycling (77) & 
MT010: Oil Analysis (68, 31.34\%), 
MT004: Refrigerant Addition (42, 19.35\%), 
MT001: Routine Maintenance (40, 18.43\%), 
MT013: Vibration Analysis (29, 13.36\%), 
MT003: Lubrication (26, 11.98\%),
MT011: Calibration, (9, 4.15\%), 
M009: Actuator Failure, (3,1.38\%) \\ \hline
\end{tabular}
\end{table}

\subsection{Insights of Work Orders Management}

\begin{figure*}[!t]
    \centering

    \begin{subfigure}[t]{0.48\textwidth}
        \centering
        \includegraphics[width=\linewidth]{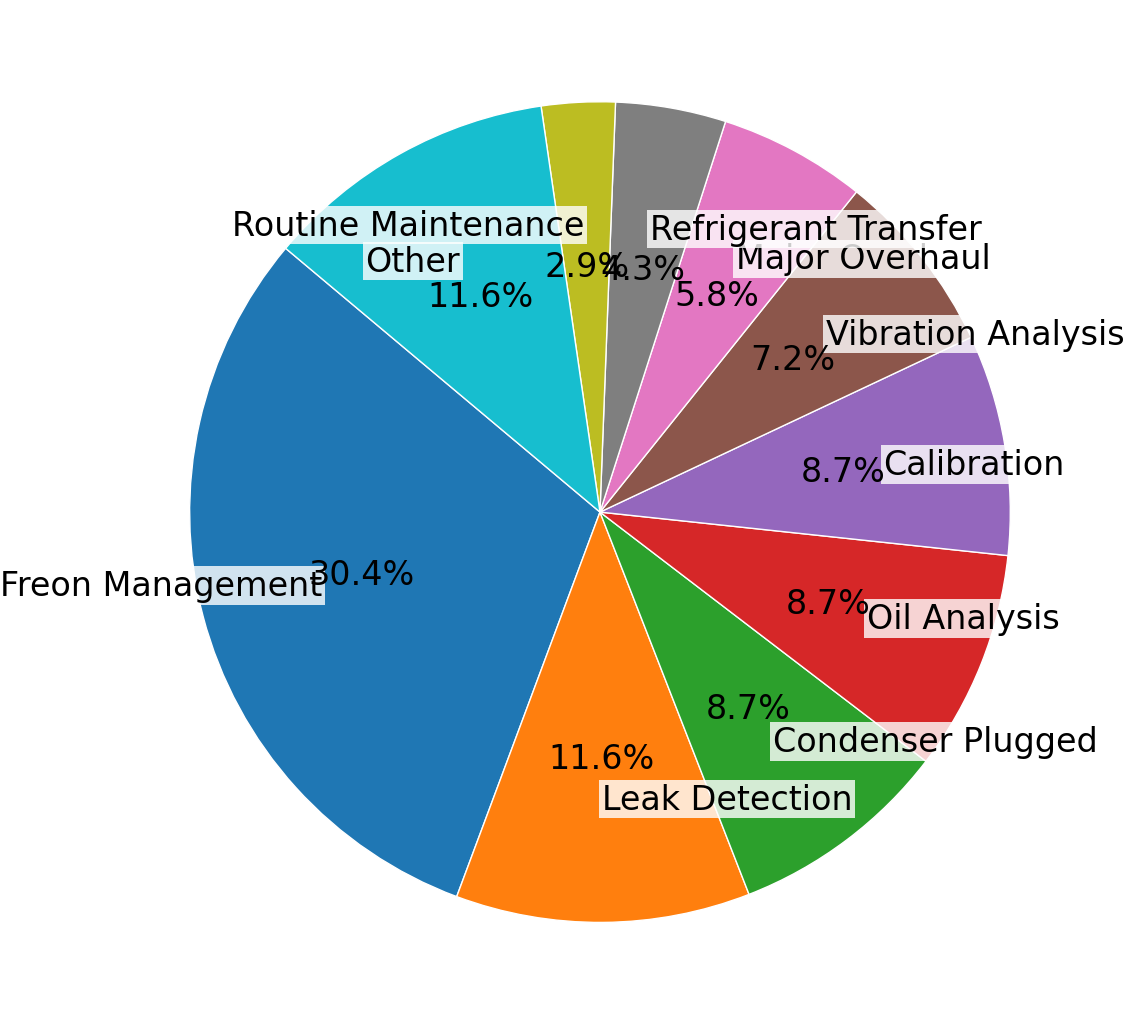}
        \caption{Distribution of WOs for Chiller 9 (Total 69 WOs)}
        \label{fig:dist_work_orde_chiller9}
    \end{subfigure}
    \hfill
    \begin{subfigure}[t]{0.48\textwidth}
        \centering
        \includegraphics[width=\linewidth]{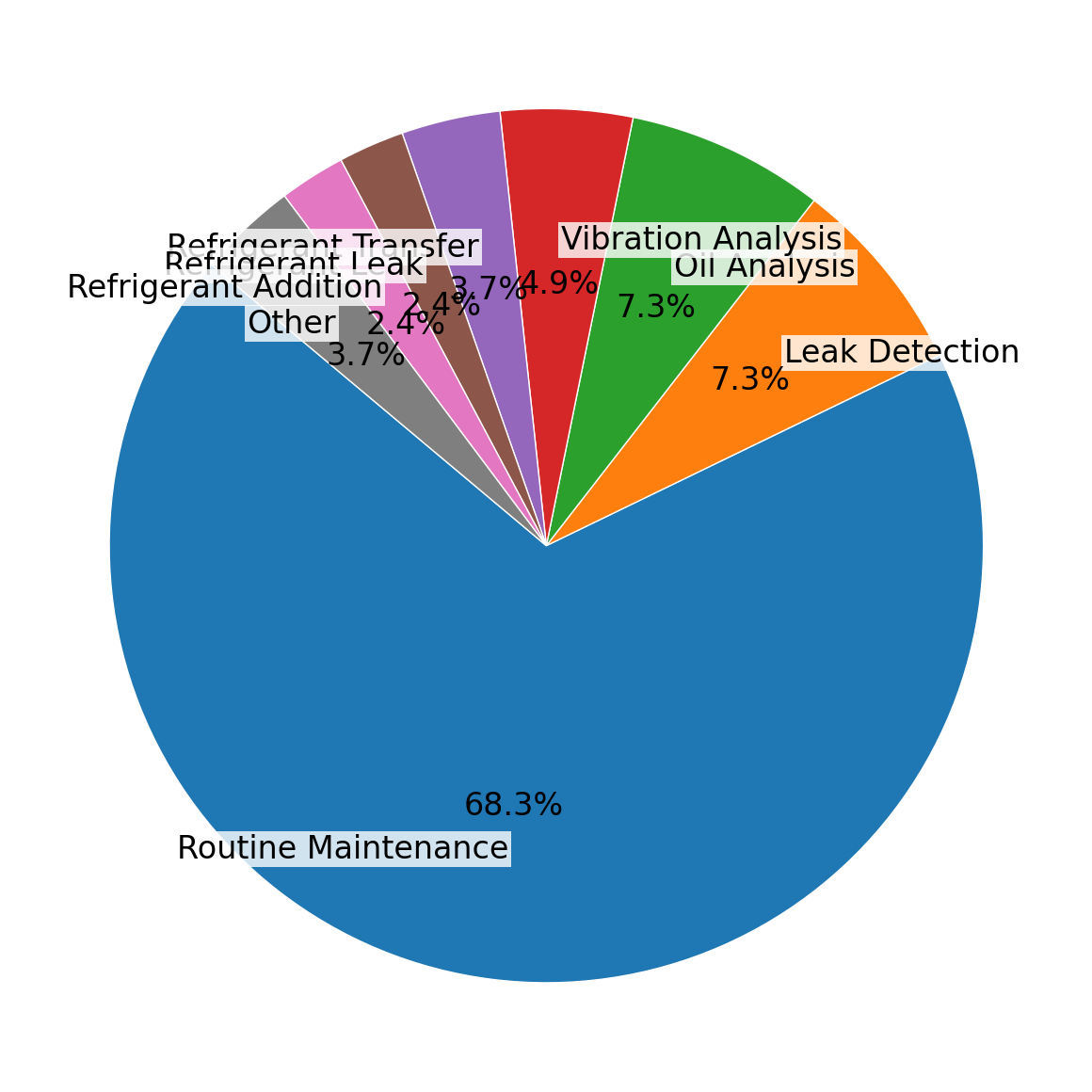}
        \caption{Distribution of WOs for Chiller 13 (Total 82 WOs)}
        \label{fig:dist_work_orde_chiller13}
    \end{subfigure}

    \par\medskip

    \begin{subfigure}[t]{0.48\textwidth}
        \centering
        \includegraphics[width=\linewidth]{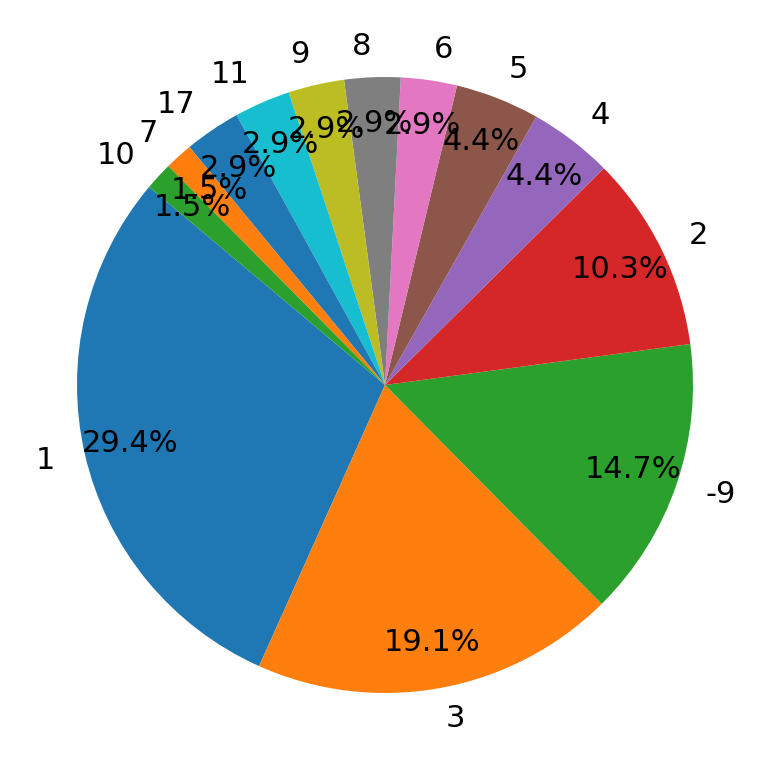}
        \caption{Matching Positions of Predicted WO Type(s)
        Versus True WO Type for Chiller 9}
        \label{fig:freq_dist_matching_positions_9}
    \end{subfigure}
    \hfill
    \begin{subfigure}[t]{0.48\textwidth}
        \centering
        \includegraphics[width=\linewidth]{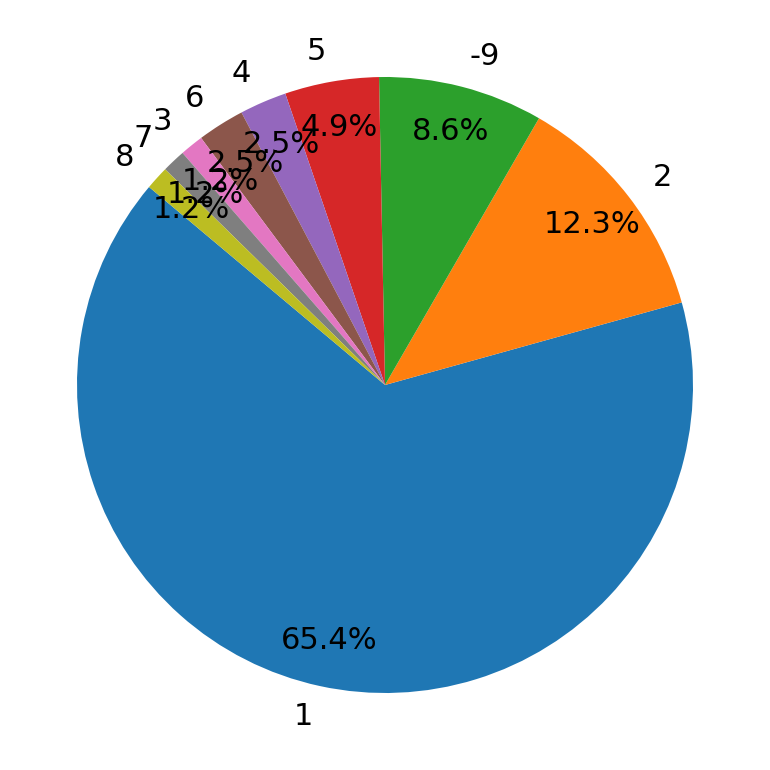}
        \caption{Matching Positions of Predicted WO Type(s)
        Versus True WO Type for Chiller 13}
        \label{fig:freq_dist_matching_positions_13}
    \end{subfigure}

    \caption{Visualization of the Distribution of Work Orders
    and Matching Positions for Next Work Order Prediction}
    \label{fig:pie_charts}
\end{figure*}

As a business case, we analyze the maintenance patterns of Chiller 9 and 13 using historical work orders and alerts to assess the framework's adaptability and predictive capabilities. Structured queries (utterances) enable the system to identify maintenance trends, predict future work orders, and evaluate the impact of alerts on corrective actions. Figures of~\ref{fig:dist_work_orde_chiller9} and~\ref{fig:dist_work_orde_chiller13} show the work order (WO) distributions for Chiller 9 and 13 and identify their different patterns.   Chiller 9 has a more varied distribution, with "Freon Management" leading at 30.4\% and several other tasks, like "Leak Detection" and "Calibration." In contrast, for Chiller 13, "Routine Maintenance" dominates WO activities at 68.3\%, with "Leak Detection" and "Oil Analysis" each at 7.3\%. These distributions highlight the emphasis on routine tasks in one scenario (Chiller 13), while the other (Chiller 9) shows a more varied maintenance approach with significant attention to multiple activities.

As shown in Figures of  \ref{fig:freq_dist_matching_positions_9} and \ref{fig:freq_dist_matching_positions_13}, we benchmarked the performance of next WO type prediction for Chiller 9 and 13. The 29.4\% (versus 65.4\%) of the predictions correctly matched the immediate next work order for Chiller 9 (versus Chiller 13), with the accuracy decreasing as the sequence progressed. Notably, 14.7\% (versus 8.6\%) of work orders did not appear within the predicted list, indicated by a match position of -9, reflecting cases where the model failed to capture the actual next work order. To generate this result, We utilized a sliding window of two years of historical WO to create a WO-specific Markov model, allowing the model to adapt to potential changes in the chiller's operation. We selected the previous two years of data for each work order in the dataset to build the Markov model, ensuring that the predictions reflected recent operational trends. The model was then used to predict the next work order type based on the current one.

\vspace{-0.2em} 

\section{Additional Implementation Details - Computing Environment and Runtime}\label{sec:addition_imp_details}
All experiments were conducted on a MacBook Pro (16-inch, 2021) with an Apple M1 Max chip, 64~GB of unified memory, and macOS Sonoma 14.4.1. Each natural language utterance was processed through multi-step agent reasoning and code generation. The inference runtime varied by task complexity: standard queries typically completed in under 30 seconds, while more complex tasks requiring iterative Python code synthesis and validation could take up to 2 minutes. To reduce unnecessary computation, we implemented an early stopping mechanism to terminate the agent's reasoning loop once valid output was produced. No large-scale training was performed, and energy usage was not directly measured, but the hardware and task scope result in moderate resource demand suitable for laptop deployment.

\end{document}